\documentclass[journal]{IEEEtran}

\ifCLASSINFOpdf
\else
\fi

\usepackage{times}
\usepackage{epsfig}
\usepackage{graphicx}
\usepackage{amsmath}
\usepackage{amssymb}
\usepackage{mathrsfs}
\usepackage{multirow}
\usepackage{blkarray}
\usepackage{float}
\usepackage{array}
\usepackage{epstopdf}
\usepackage{subcaption}
\usepackage[T1]{fontenc}
\usepackage{algorithm}
\usepackage{algpseudocode}

\begin{document}

\title{Subspace Learning in The Presence of Sparse Structured Outliers and Noise}

\author{Shervin~Minaee, and~Yao~Wang
}

\maketitle

\begin{abstract}
Subspace learning is an important problem, which has many applications in image and video processing.
It can be used to find a low-dimensional representation of signals and images. 
But in many applications, the desired signal is heavily distorted by outliers and noise, which negatively affect the learned subspace.
In this work, we present a novel algorithm for learning a subspace for signal representation, in the presence of structured outliers and noise.
The proposed algorithm tries to jointly detect the outliers and learn the subspace for images.
We present an alternating optimization algorithm for solving this problem, which iterates between learning the subspace and finding the outliers.
This algorithm has been trained on a large number of image patches, and the learned subspace is used for image segmentation, and is shown to achieve better segmentation results than prior methods, including least absolute deviation fitting, k-means clustering based segmentation in DjVu, and shape primitive extraction and coding algorithm.
\end{abstract}


\IEEEpeerreviewmaketitle

\section{Introduction}
Many of the signal and image processing problems can be posed as problems of learning a low dimensional linear or multi-linear model.
Algorithms for learning linear models can be seen as a special case of subspace fitting. Many of these algorithms are based on least squares estimation techniques, such as principal component analysis (PCA) \cite{pca}, linear discriminant analysis (LDA) \cite{lda}, and locality preserving projection \cite{lpp}.
But in general, training data may contain undesirable artifacts due to occlusion, illumination changes, overlaying component (such as foreground texts and graphics on top of smooth background image). These artifacts can be seen as outliers for the desired signal.
As it is known from statistical analysis, algorithms based on least square fitting fail to find the underlying representation of the signal in the presence of outliers \cite{lsf}.
Different algorithms have been proposed for robust subspace learning to handle outliers in the past, such as the work by Torre \cite{rsl}, where he suggested an algorithm based on robust M-estimator for subspace learning.
Robust principal component analysis \cite{rpca} is another approach to handle the outliers.
In \cite{lerman}, Lerman et al proposed an approach for robust linear model fitting by parameterizing linear subspace using orthogonal projectors.
There have also been many works for online subspace learning/tracking for video background subtraction, such as GRASTA \cite{grasta}, which uses a robust $\ell_1$-norm cost function in order to estimate and track
non-stationary subspaces when the streaming data vectors are corrupted with outliers, and t-GRASTA \cite{tgrasta}, which simultaneously estimate a decomposition of a collection of images into a low-rank subspace, and sparse part, and a transformation such as rotation or translation of the image.

In this work, we present an algorithm for subspace learning from a set of images, in the presence of structured outliers and noise.
We assume some structure on outliers that suits many of the image processing applications, which is connectivity and sparsity. 
As a simple example we can think of smooth images overlaid with texts and graphics foreground, or face images with occlusion (as outliers).
To promote the connectivity of the outlier component, the group-sparsity \cite{group} of outlier pixels is added to the cost function (It is worth mentioning that total-variation \cite{tv} can also be used to promote connectivity). 
We also impose the smoothness prior on the learned subspace representation, by penalizing the gradient of the representation.
We then propose an algorithm based on the sparse decomposition framework for subspace learning. This algorithm jointly detect the outlier pixels and learn the low-dimensional subspace for underlying image representation. 

After learning the subspace, we present its application for background-foreground segmentation in still images, and show that it achieves better performance than previous algorithms.
We compare our algorithm with some of the prior approaches, including k-means clustering in DjVu \cite{djvu}, shape primitive extraction and coding (SPEC) \cite{spec}, least absolute deviation fitting (LAD) \cite{LAD}.
The proposed algorithm has applications in text extraction, medical image analysis, and image decomposition \cite{mine1}-\cite{mine6}.

One problem with previous clustering-based segmentation techniques is that if the intensity of background pixels has a large dynamic range, some part of the background could be segmented as foreground, but our proposed model can correctly segment the image.
One such example is shown in Fig. 1, where the foreground mask (a binary mask showing the location of foreground pixels) for a sample image by clustering and our algorithm are shown.
\begin{figure}
\begin{center}
\hspace{-0.15cm}
    \includegraphics [scale=0.44] {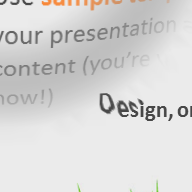}
  \hspace{0.11cm}  \includegraphics [scale=0.26] {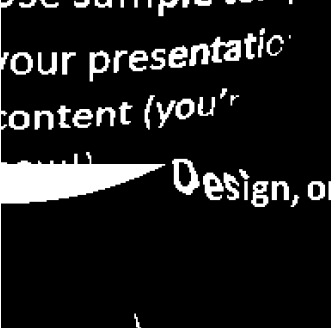} 
  \hspace{-0.29cm}  \includegraphics [scale=0.15] {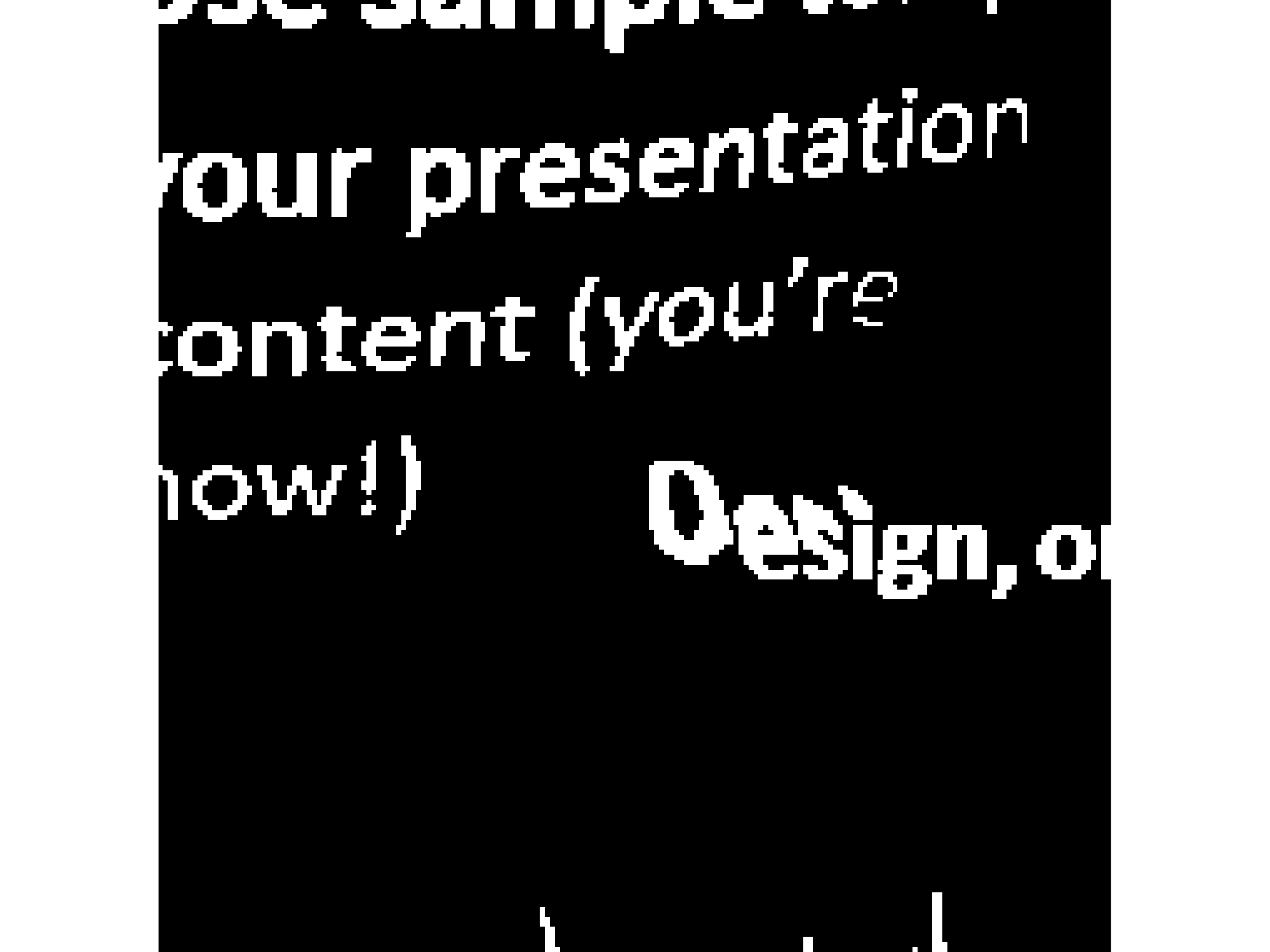} 
\end{center}
  \caption{The left, middle and right images denote the original image, segmented foreground by hierarchical k-means and the proposed algorithm respectively.}
\end{figure}
\vspace{-0.1cm}

The structure of the rest of this paper is as follows: Section II presents the proposed framework for subspace learning. 
The detail of alternating optimization problem is presented in Section II. A, and the application of this framework for image segmentation is presented in II. B.
Section III provides the experimental results for the proposed algorithm and its application for image segmentation.
And finally the paper is concluded in Section IV.

\section{The Problem Formulation}
Despite the high-dimensionality of images (and other kind of signals), many of them have a low-dimensional representation.
For some category of images, this low-dimensional representation may be a very complex manifold which is not simple to find, but for many of the smooth images this low-dimensional representation can be assumed to be a subspace. 
Therefore each signal $x \in R^N$ can be efficiently represented as:
\begin{equation}
\begin{aligned}
x\simeq P\alpha
\end{aligned}
\end{equation}
where  $P \in R^{N \times k}$ where $k \ll N$, and $\alpha$ denotes the representation coefficient in the subspace.
\\There have been many approaches in the past to learn $P$ efficiently, such as PCA and robust-PCA.
But in many scenarios, the desired signal can be heavily distorted with outliers and noise, and those distorted pixels should not be taken into account in subspace learning process, since they are assumed to not lie on the desired signal subspace.
Therefore a more realistic model for the distorted signals should be:
\begin{equation}
\begin{aligned}
x= P \alpha+s+\epsilon
\end{aligned}
\end{equation}
where $s$ and $\epsilon$ denote the outlier and noise components respectively.
Here we propose an algorithm to learn a subspace, $P$, from a training set of $N_d$ samples $x_i$, by minimizing the noise energy ($\|\epsilon_i  \|_2^2= \|x_i-P\alpha_i-s_i  \|_2^2$), and some regualrization term on each component, as:
\begin{equation}
\begin{aligned}
& \underset{P, \alpha_i, s_i}{\text{min}}
 \sum_{i=1}^{N_d} \ \frac{1}{2} \| x_i-P\alpha_i-s_i  \|_2^2+ \lambda_1  \phi(P\alpha_i) + \lambda_2 \psi(s_i)  \\
& \ \text{s.t.}
\ \ \ \ \ \ \ \ P^tP= I, \ s_i \geq 0
\end{aligned}
\end{equation}
where $\phi(.)$ and $\psi(.)$ denote suitable regularization terms on the first and second components, promoting our prior knowledge about them.
Here we assume the underlying image component is smooth, therefore it should have a small gradient. And for the outlier, we assume it is sparse and also connected, therefore we need to promote the sparsity and connectivity \cite{mairal}.
Hence $\phi(P\alpha_i)= \| \nabla P \alpha_i \|_2^2$, and $\psi(s)= \|s\|_1+ \beta \sum_m \|s_{g_m}\|_2$, where $g_m$ shows the m-th group in the outlier (the pixels within each group are supposed to be connected).
\\Putting all these together, we will get the following optimization problem:
\begingroup\makeatletter\def\f@size{9}\check@mathfonts
\begin{equation}
\begin{aligned}
& \hspace{-1cm } \underset{P, \alpha_i, s_i}{\text{min}}
  \sum_{i=1}^{N_d} \frac{1}{2} \| x_i-P\alpha_i-s_i  \|_2^2+ \lambda_1  \| \nabla P \alpha_i \|_2^2 + \lambda_2 \|s_i\|_1+ \lambda_3 \sum_m \|s_{i,g_m}\|_2  \\
& \ \text{s.t.}
\ \ \ \ \ \ \ \ P^tP= I, \ s_i \geq 0
\end{aligned}
\end{equation}
\endgroup
Here by $s_i \geq 0$ we mean all elements of the vector $s_i$ should be non-negative.
Note that $ \| \nabla P \alpha_i \|_2^2$ denotes the spatial gradient, which can be written as:
\begin{equation}
\begin{aligned}
\| \nabla P \alpha_i \|_2^2= \| D_x P \alpha_i \|_2^2+ \| D_y P \alpha_i \|_2^2= \| D P \alpha_i \|_2^2
\end{aligned}
\end{equation}
where $D_x$ and $D_y$ denote the horizontal and vertical derivative matrix operators, and $D=[D_x^t, D_y^t]^t$.

\subsection{The Alternating Optimization Approach}
The optimization problem in Eq (4) can be solved using alternating optimization over $\alpha_i$, $s_i$ and $P$.
In the following part, we present the update rule for each variable by setting the gradient of cost function w.r.t that variable to zero.
\\The update step for $\alpha_i$ would be:
\begin{small}
\begin{equation*}
\begin{aligned}
\alpha_i^*= & \underset{ \alpha_i}{\text{\ \ argmin}}
 \{ \frac{1}{2} \| x_i- P\alpha_i- s_i  \|_2^2+ \frac{\lambda_1}{2} \| D P \alpha_i \|_2^2 = F_{\alpha}(\alpha_i) \} \Rightarrow \\
& \hspace{-0.72cm}  \nabla_{\alpha_i}F_{\alpha}(\alpha_i^*)=0 \Rightarrow P^t(P\alpha_i^*+ s_i- x_i)+ \lambda_1 P^t D^t D P \alpha_i^*=0 \Rightarrow \\
& \hspace{-0.66cm} \alpha_i^*= (P^t P+ \lambda_1 P^t D^t D P )^{-1} P^t (x_i-s_i) \\
\end{aligned}
\end{equation*}
\end{small}

The update step for the $m$-th group of the variable $s_i$ is as follows:
\begin{small}
\begin{equation*}
\begin{aligned}
& \  s_{i,g_m}= \underset{ s_i}{\text{\ \ argmin}}
 \{ \frac{1}{2} \| (x_i- P\alpha_i)_{g_m}- s_{i,g_m}  \|_2^2+  \lambda_2 \| s_{i,g_m} \|_1+ \\ 
& \ \lambda_3 \|s_{i,g_m}\|_2 = F_s(s_{i,g_m})\} \ \ \ \ \ \ \ \ \ \ \ \text{s.t.} \ \ \ \ \ \ s_{i,g_m} \geq 0 \\ 
&  \Rightarrow \nabla_{s_{i,g_m}}F_{s}(s_{i,g_m})=0 \Rightarrow s_{i,g_m}+(P\alpha_i-x_i)_{g_m}+ \lambda_2 \text{sign}(s_{i,g_m}) \\ 
& +\lambda_3 \frac{s_{i,g_m}}{\ \|s_{i,g_m}\|_2}= 0 \Rightarrow s_{i,g_m}+ \lambda_3 \frac{s_{i,g_m}}{\ \|s_{i,g_m}\|_2}= (x_i-P\alpha_i)_{g_m}-\lambda_2 1\\
& \ \hspace{-0.1cm} \Rightarrow s_{i,g_m}= \text{block-soft} ((x_i-P\alpha_i)_{g_m}-\lambda_2 1, \ \lambda_3)  \\
\end{aligned}
\end{equation*}
\end{small}
Note that, because of the constraint $s_{i,g_m} \geq 0$, we can approximate $\text{sign}(s_{i,g_m})=1$, and then project the $s_{i,g_m}$ from soft-thresholding result onto $s_{i,g_m} \geq 0$, by setting its negative elements to 0.   
The block-soft(.) \cite{admm} is defined as:
\begin{equation*}
\begin{aligned}
\text{block-soft}(y,t)= \text{max}(1-\frac{t}{\ \|y\|_2},0) \ y
\end{aligned}
\end{equation*}
\\For the subspace update, we first ignore the orthonormality constraint ($P^tP=I$), and update the subspace column by column, and then use Gram-Schmidt algorithm \cite{gram} to orthonormalize the columns. If we denote the j-th column of $P$ by $p_j$, its update can be derived as:
\begin{small}
\begin{equation*}
\begin{aligned}
& \ P= \underset{ P}{\text{\ \ argmin}}
 \{\sum_i \frac{1}{2} \| x_i- P\alpha_i- s_i  \|_2^2+  \lambda_1 \| D P \alpha_i \|_2^2 \} \Rightarrow \\ 
& \ p_j= \underset{ p_j}{\text{\ \ argmin}}
 \{\sum_i \frac{1}{2} \| (x_i- \sum_{k \ne j}p_k\alpha_i(k)- s_i) - p_j \alpha_i(j) \|_2^2+ \\
 & \ \lambda_1 \| D \sum_{k \neq j} p_k \alpha_i(k)+ D p_j \alpha_i(j) \|_2^2 = \sum_i  \frac{1}{2} \| \eta_{i,j} - p_j \alpha_i(j) \|_2^2+ \\
 &  \ \lambda_1 \| \gamma_{i,j}+ D p_j \alpha_i(j) \|_2^2 = F_{p}(p_j) \} \Rightarrow \nabla_{p_j}F_{p}(p_j^*)=0 \Rightarrow \\
 &  \ \sum_i \alpha_i(j) \big(\alpha_i(j)p_j-\eta_{i,j}\big)+ \lambda_1 \alpha_i(j) D^t \big( \alpha_i(j) Dp_j + \gamma_{i,j}  \big)=0 \Rightarrow \\
  & \  \big( \sum_i \alpha_i^2(j) \big) (I+ \lambda_1 D^tD) p_j= \sum_i \big( \alpha_i(j) \eta_{i,j}-\lambda_1 \alpha_i(j)D^t \gamma_{i,j} \big)= \beta_j \\
  & \ \Rightarrow p_j= (I+ \lambda_1 D^tD)^{-1} \beta_j /\big( \sum_i \alpha_i^2(j) \big) \\
\end{aligned}
\end{equation*}
\end{small}
where $\eta_{i,j}=x_i-s_i- \sum_{k \ne j}p_k\alpha_i(k)$, and $\gamma_{i,j}=D \sum_{k \neq j} p_k \alpha_i(k)$.
After updating all columns of $P$, we apply Gram-Schmidt algorithm to project the learnt subspace onto $P^tP=I$. Note that orthonormalization should be done at each step of alternating optimization.
It is worth to mention that for some applications the non-negativity assumption for the structured outlier may not be valid, so in those cases we will not have the $s_i \geq 0$ constraint. In that case, the problem can be solved in a similar manner, but we need to introduce an auxiliary variable $s=z$, to be able to get a simple update for each variable.

\subsection{Applications For Image Segmentation}
After learning the subspace, it can be used for different applications, such as segmentation and classification of signals.
Here we use this framework for background-foreground segmentation in still images. 
Suppose we want to separate the foreground texts and graphics from background regions. We can think of foreground as the outliers overlaid on top of background, and use the learned subspace along with the following sparse decomposition framework to separate them:
\begin{equation}
\begin{aligned}
& \underset{\alpha, s}{\text{min}}
  \frac{1}{2} \| x-P\alpha-s  \|_2^2+ \lambda_1  \| D P \alpha \|_2^2 + \lambda_2 \|s\|_1+ \lambda_3 \sum_m \|s_{g_m}\|_2    \\
& \ \text{s.t.}
\ \ \ \ \ \ \ \ s \geq 0
\end{aligned}
\end{equation}
In our image segmentation problem, the m-th column of each block is chosen as the m-th group, $g_m$. 
The reason being there are more vertical connectivity in English texts than horizontal. 
We could also impose both column-wise and row-wise connectivity, but it would require introducing auxiliary variables in the optimization framework.
The problem in (6) can be easily solved using ADMM \cite{admm}, and proximal optimization \cite{prox}.
After solving this problem, the $s$ component will be thresholded to find the foreground position.

\section{Experimental Results}
To evaluate the performance of our algorithm, we trained the proposed framework on image patches extracted from some of the images of the screen content image segmentation dataset provided in \cite{LAD}. 
Before showing the results, we will report the weight parameters in our optimization.
We used $\lambda_1=0.5$, $\lambda_2=1$ and $\lambda_3=2$, which are tuned by testing on a validation set.
We provide the results for subspace learning and image segmentation in the following sections.

\subsection{The Learned Subspace}
We extracted around 8,000 overlapping patches of size 32x32, with stride of 5 from a subset of these images and used them for learning the subspace, and learned a 64 dimensional subspace (which means 64 basis images of size 32x32). 
The learned atoms of this subspace are shown in Figure 2.

\begin{figure}
\begin{center}
    \includegraphics [scale=0.38] {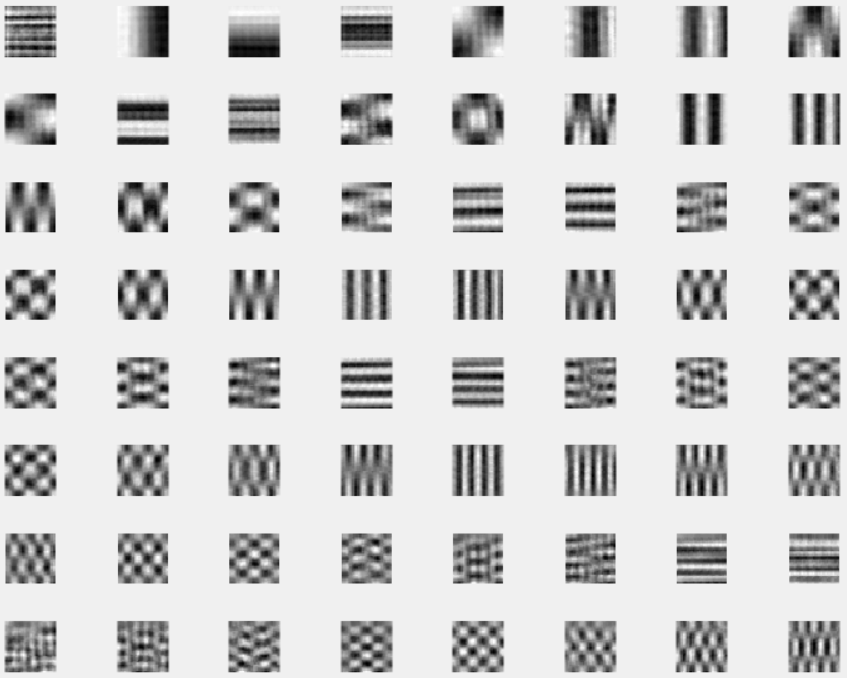}
  \caption{The learned 64 basis images (a subspace of dimension 64) for 32x32 image blocks}
\end{center}
\end{figure}

As we can see the learned atoms contain different edge and texture patterns, which is reasonable for image representation. The right value of subspace dimension highly depends to the application. 
For image segmentation problem studied in this paper, we found that using only first 20 atoms performs well on image patches of 32x32.
The experiments are performed using MATLAB 2015 on a laptop with Core i5 CPU running at 2.2GHz. 
It takes around 78 seconds to learn the 64 dimensional subspace.

\subsection{Applications in Image Segmentation}
After learning the subspace, we use this representation for background-foreground segmentation in still images, as explained in Section II.B. 
The segmentation results in this section are derived by using a 20 dimensional subspace for background modeling.
We use the same model as the one in Eq (6) for decomposition of an image into background and foreground, and $\lambda_i$'s are set to the same value as mentioned before.
We then evaluate the performance of this algorithm on the remaining images from screen content image segmentation dataset \cite{our_dataset}, and some other images, and compare the results  with three previous algorithms; hierarchical k-means clustering in DjVu \cite{djvu}, SPEC \cite{spec}, sparse and low-rank decomposition \cite{lowrank}, and LAD \cite{LAD}.
For sparse and low rank decomposition, we apply the fast-RPCA algorithm \cite{lowrank} on the image blocks, and threshold the sparse component to find the foreground location.
For low-rank decomposition, we have used the MATLAB implementation provided by Stephen Becker at \cite{becker}.

To provide a numerical comparison, we report the average precision, recall and F1 score \cite{metrics} achieved by different algorithms over this dataset. The average precision, recall and F1 score by different algorithms are given in Table 1. 

\begin{table}[ht]
\centering
  \caption{Comparison of accuracy of different algorithms}
  \centering
\begin{tabular}{|m{3.7cm}|m{1.1cm}|m{1cm}|m{1.1cm}|}
\hline
Segmentation Algorithm  &  Precision & \ \  Recall & \  F1 score\\
\hline
SPEC \cite{spec} & \ \ \ 50\% & \ \ \  64\% & \ \ \ 56.1\% \\
\hline
 Hierarchical Clustering \cite{djvu} & \ \ \ 64\% & \ \ \ 69\% & \ \ \ 66.4\% \\
\hline 
 Low-rank Decomposition  \cite{lowrank} & \ \ \  78\% & \ \ \  86.5\% & \ \ \  82.1\% \\
\hline 
 Least Absolute Deviation \cite{LAD} & \ \ \  91.4\% & \ \ \  87\% & \ \ \  89.1\% \\
\hline
 The proposed algorithm & \ \ \ 93\%  & \ \ \ 86\%  & \ \ \  89.3\%\\
\hline
\end{tabular}
\label{TblComp}
\end{table}

The precision and recall are defined as in Eq. (7), where TP, FP and FN denote true positive, false positive and false negative respectively. In our evaluation, we treat a foreground pixel as positive.
The balanced F1 score is defined as the harmonic mean of precision and recall, as it is shown in Eq 8.
\begin{gather}
 \text{Precision}= \frac{\text{TP}}{\text{TP+FP}} \ , 
\ \ \ \ \text{Recall}= \frac{\text{TP}}{\text{TP+FN}} 
\end{gather}
\begin{gather}
\text{F1}= 2 \ \frac{\text{precision} \times \text{recall}}{\text{precision+recall}}
\end{gather}
As it can be seen, the proposed scheme achieves much higher precision and recall than hierarchical k-means clustering and SPEC algorithms. Compared to the least absolute deviation fitting, the proposed formulation has slightly better performance.

\begin{figure}
        \centering
        \hspace{-1.2cm}
        \begin{subfigure}[b]{0.18\textwidth}
       \vspace{-3cm}
                \includegraphics[width=\textwidth]{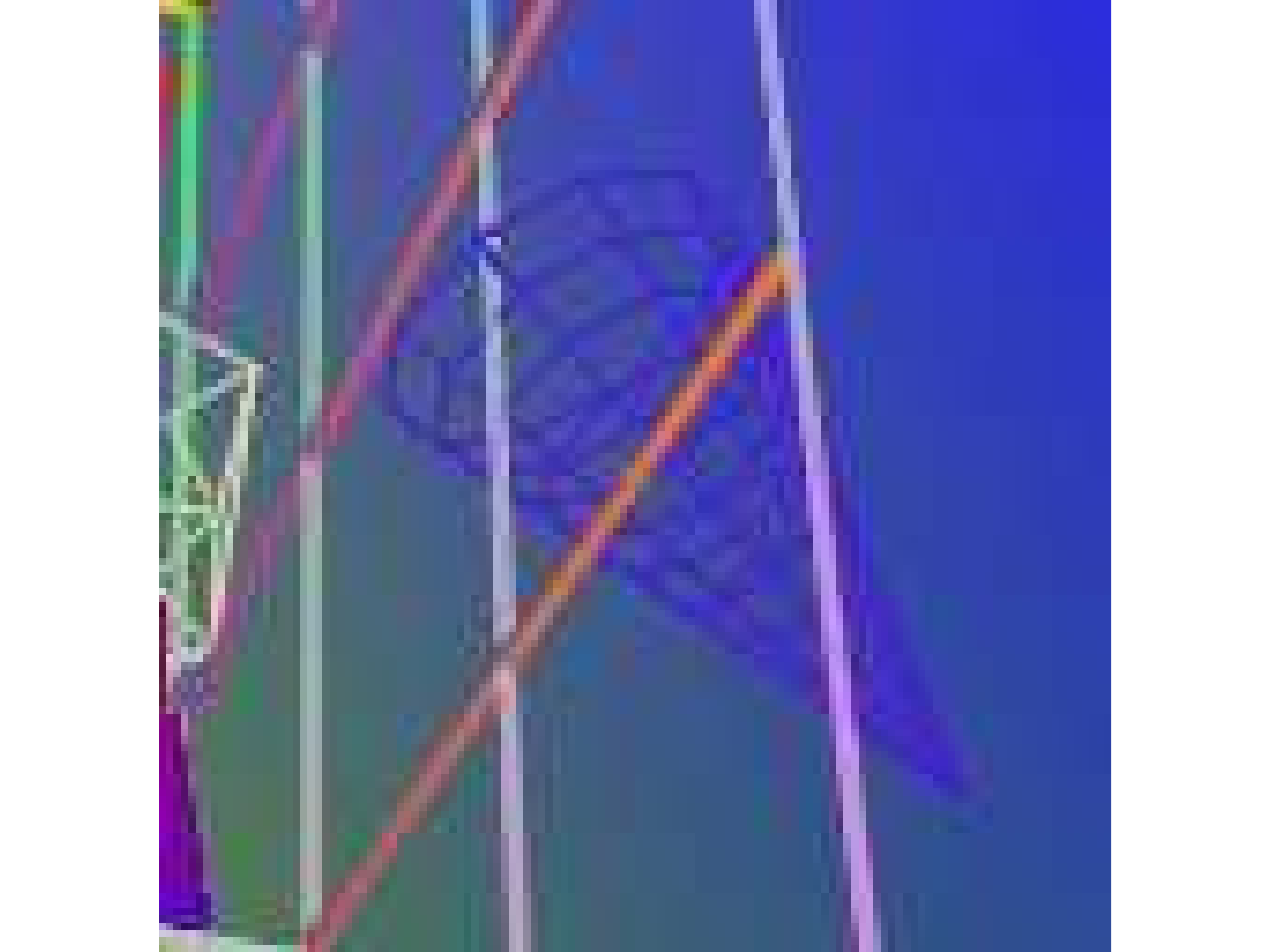}
                                \vspace{-0.35cm}
          \hspace{-0.92cm}    
        \end{subfigure}%
        ~ 
        \begin{subfigure}[b]{0.18\textwidth}
       \hspace{-0.5cm}
       \vspace{0.17cm}
                \includegraphics[width=\textwidth]{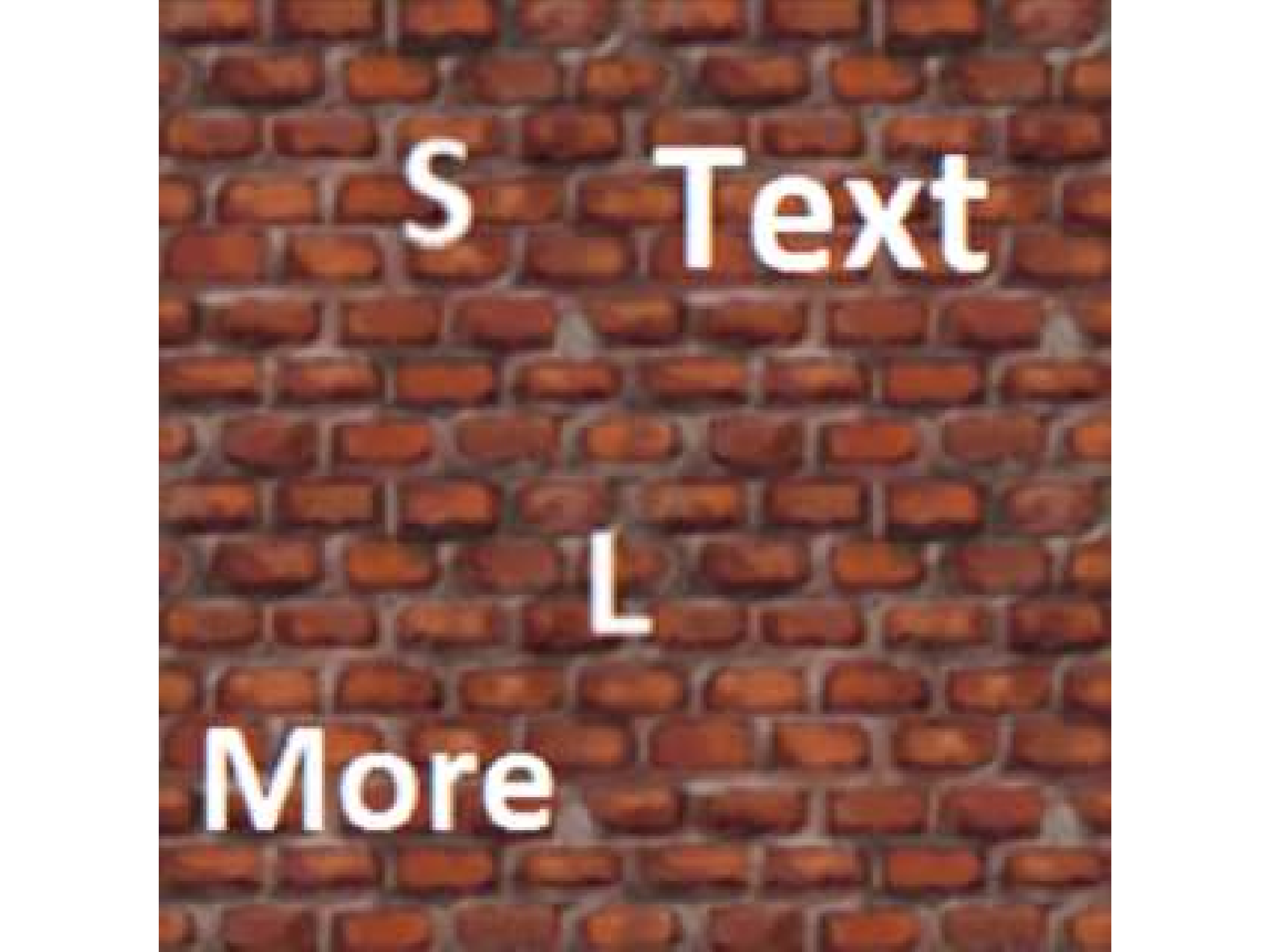}
                \vspace{-0.5cm}
            \hspace{-1.5cm} 
        \end{subfigure}%
        ~ 
        \begin{subfigure}[b]{0.18\textwidth}
              \hspace{-0.95cm}
       \vspace{0.13cm}
                \includegraphics[width=\textwidth]{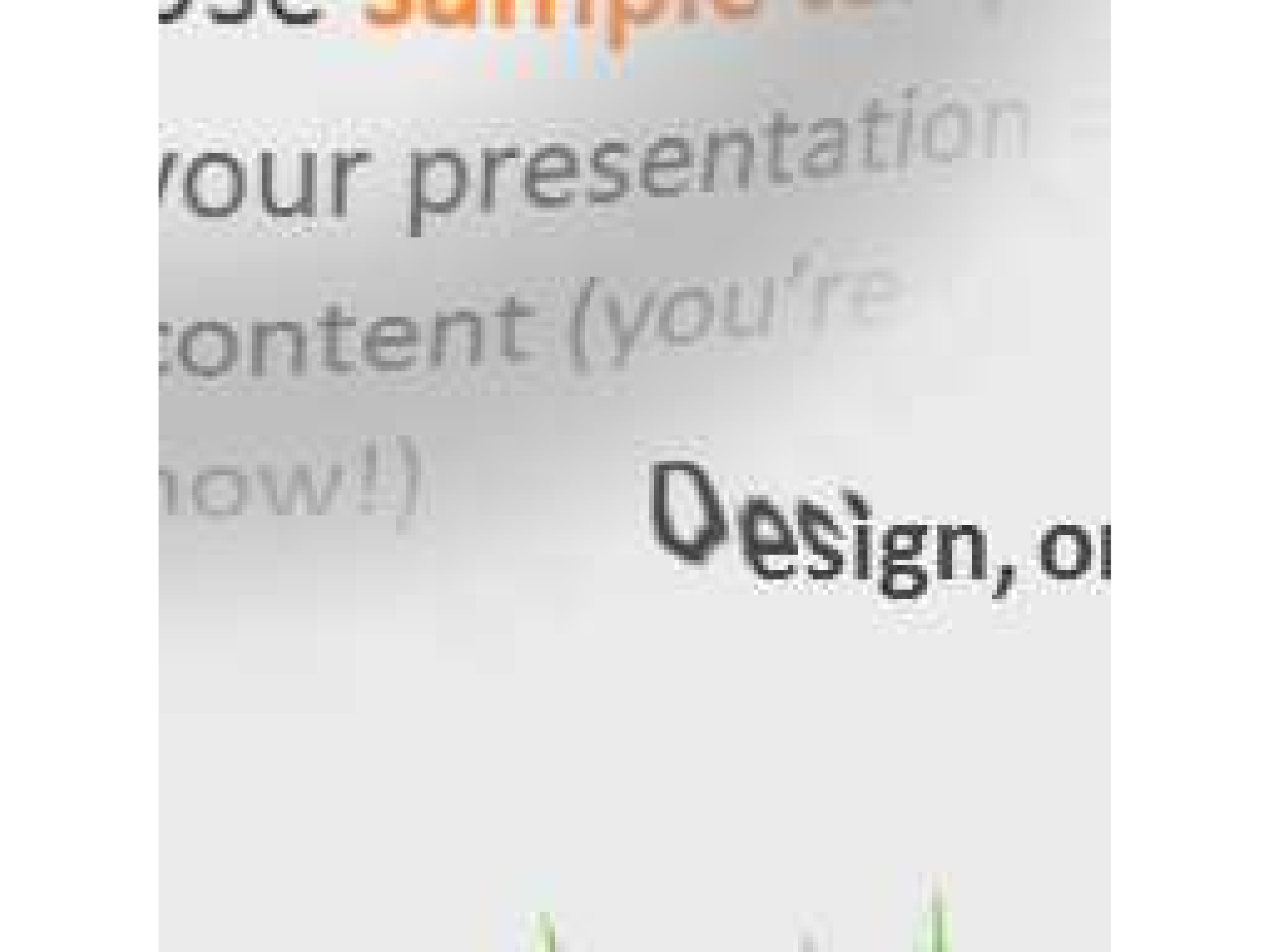}
                 \vspace{-0.45cm}
              \hspace{-1.8cm}
        \end{subfigure}
         \\[1ex]
                 \centering
        \vspace{-0.05cm}
        \hspace{-1.2cm}
        \begin{subfigure}[b]{0.18\textwidth}
                \includegraphics[width=\textwidth]{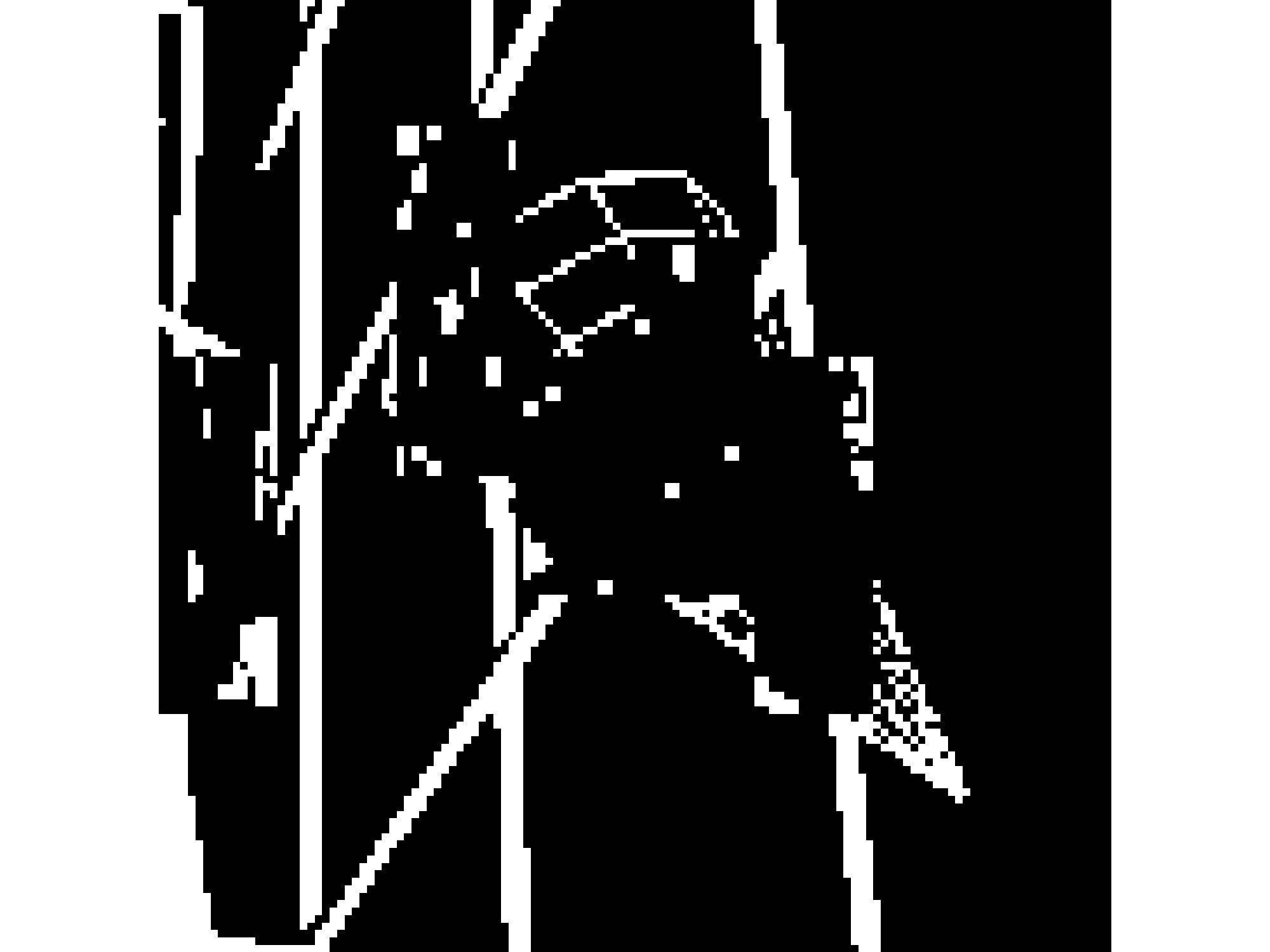}
                                \vspace{-0.35cm}
          \hspace{-2.5cm}    
        \end{subfigure}%
        ~ 
        \begin{subfigure}[b]{0.18\textwidth}
              \hspace{-0.5cm}
       \vspace{0.17cm}
                \includegraphics[width=\textwidth]{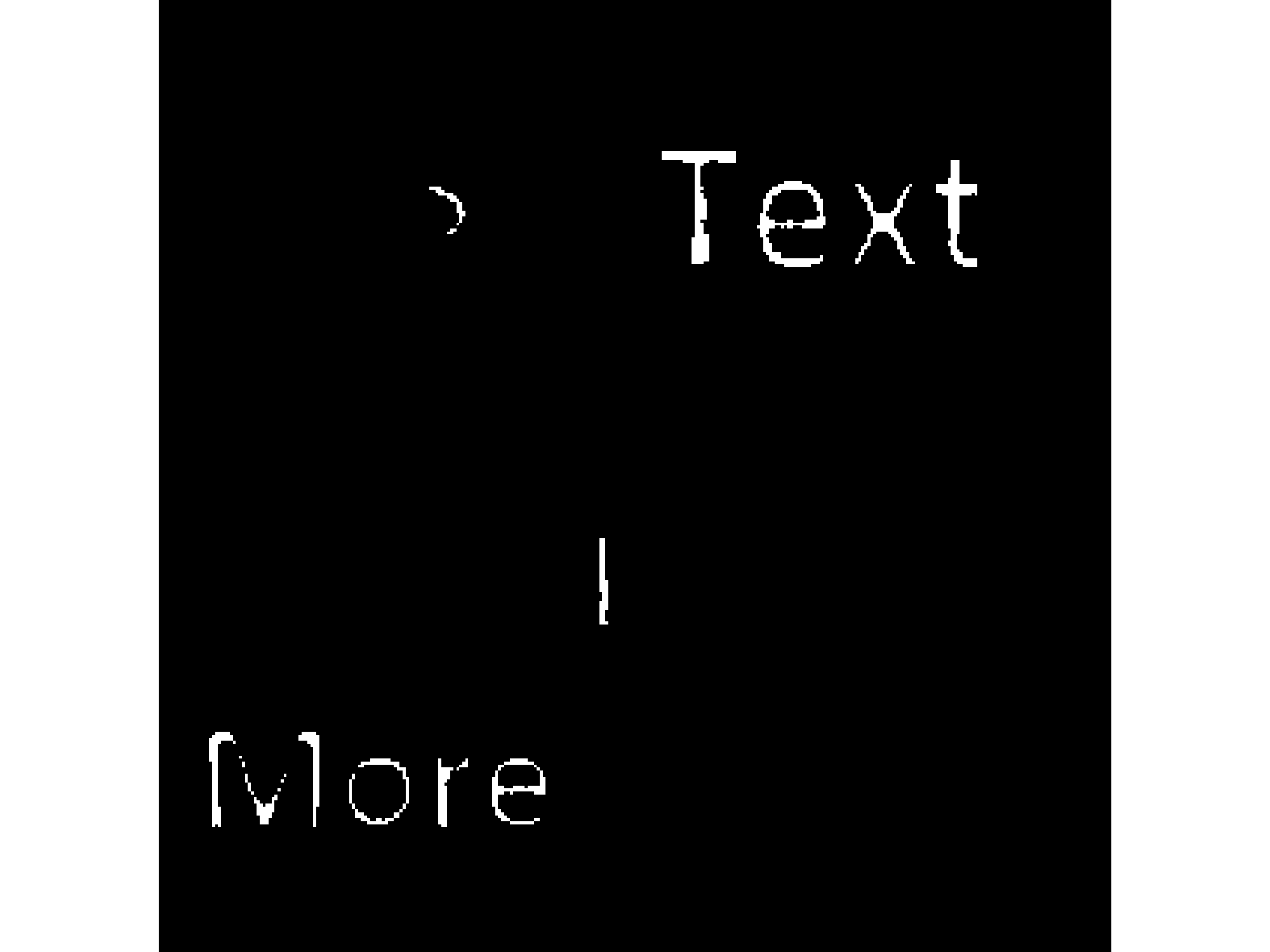}
                \vspace{-0.5cm}
            \hspace{-3cm} 
        \end{subfigure}%
        ~ 
        \begin{subfigure}[b]{0.18\textwidth}
              \hspace{-0.995cm}
       \vspace{0.13cm}
                \includegraphics[width=\textwidth]{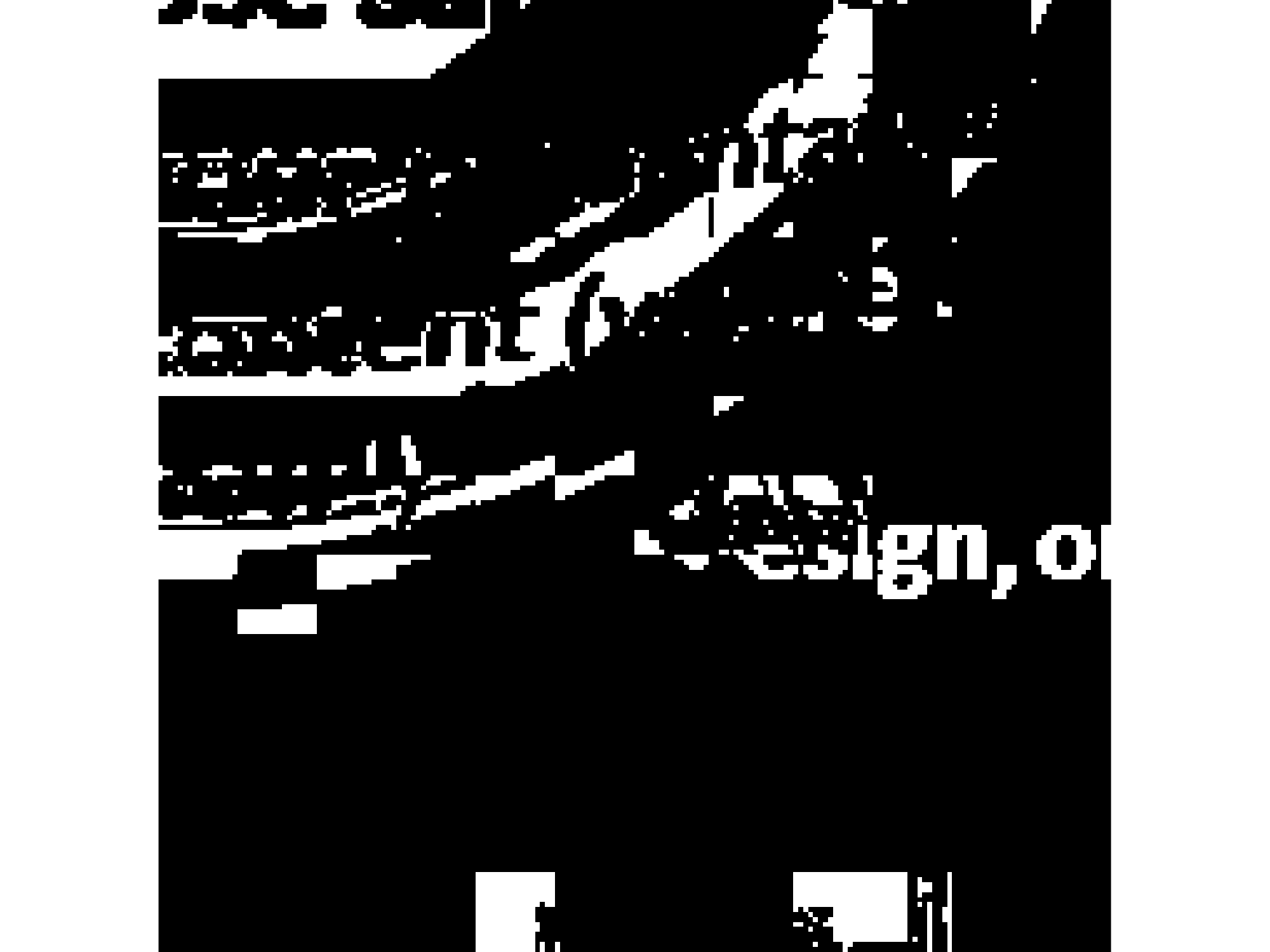}
                 \vspace{-0.45cm}
              \hspace{-4.8cm}
        \end{subfigure}
        \\[1ex]
                 \centering
        \vspace{-0.05cm}
        \hspace{-1.2cm}
        \begin{subfigure}[b]{0.18\textwidth}
                \includegraphics[width=\textwidth]{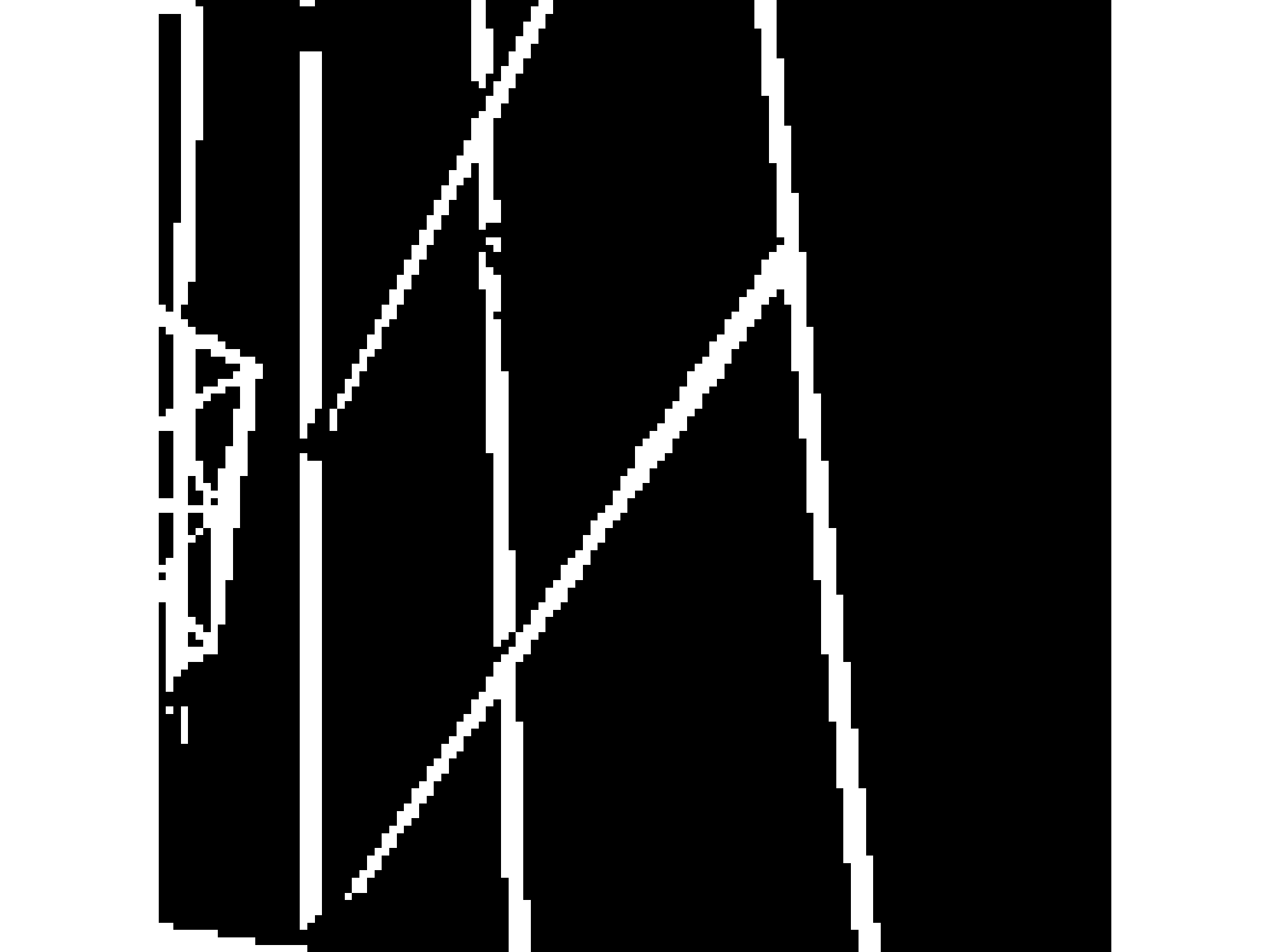}
                                \vspace{-0.35cm}
          \hspace{-2.5cm}    
        \end{subfigure}%
        ~ 
        \begin{subfigure}[b]{0.18\textwidth}
               \hspace{-0.5cm}
       \vspace{0.17cm}
                \includegraphics[width=\textwidth]{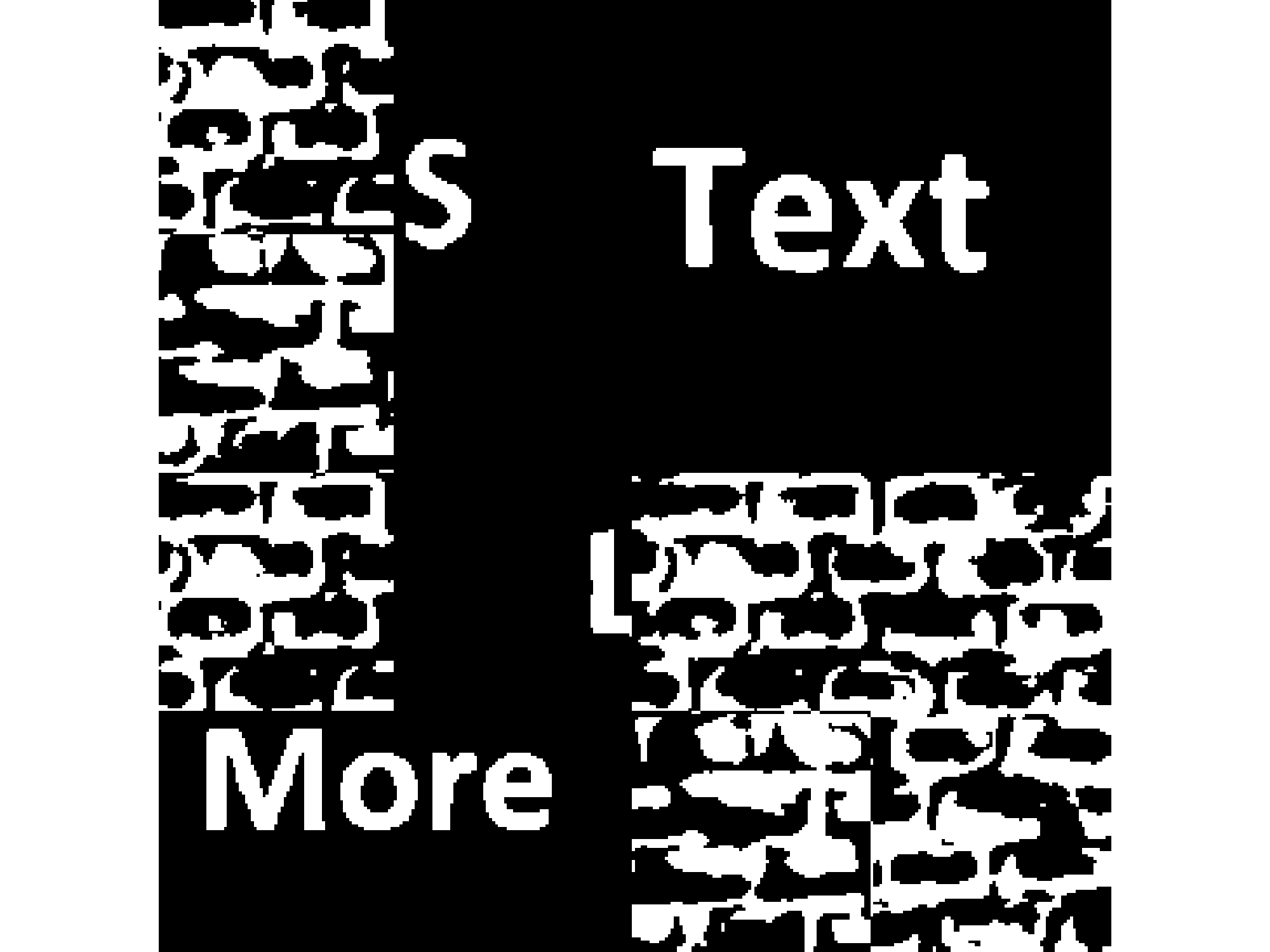}
                \vspace{-0.5cm}
            \hspace{-3cm} 
        \end{subfigure}%
        ~ 
        \begin{subfigure}[b]{0.18\textwidth}
               \hspace{-0.995cm}
       \vspace{0.13cm}
                \includegraphics[width=\textwidth]{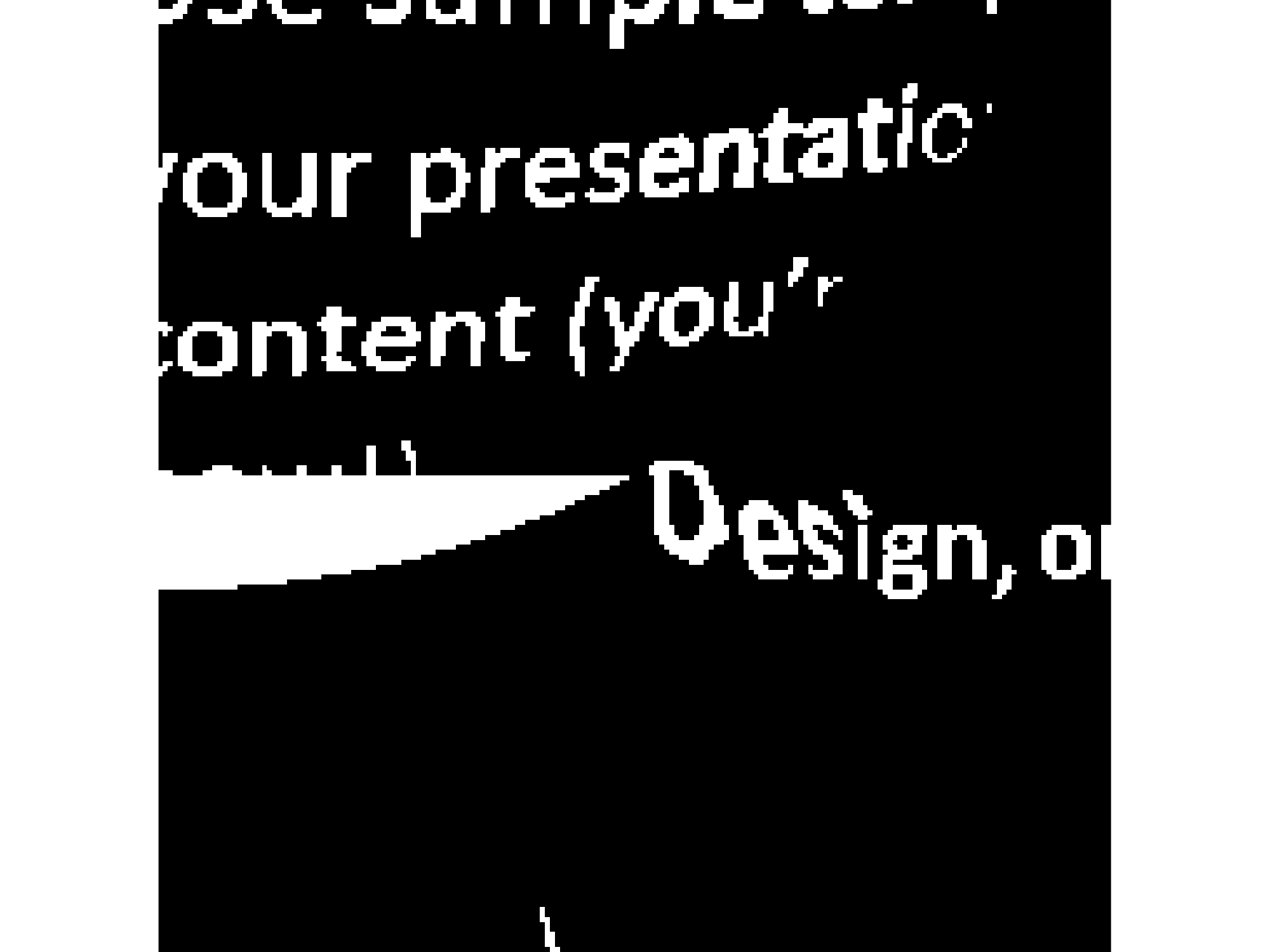}
                 \vspace{-0.45cm}
              \hspace{-4.8cm}
        \end{subfigure}
        \\[1ex]         
                 \centering
        \vspace{-0.05cm}
        \hspace{-1.2cm}
        \begin{subfigure}[b]{0.18\textwidth}
                \includegraphics[width=\textwidth]{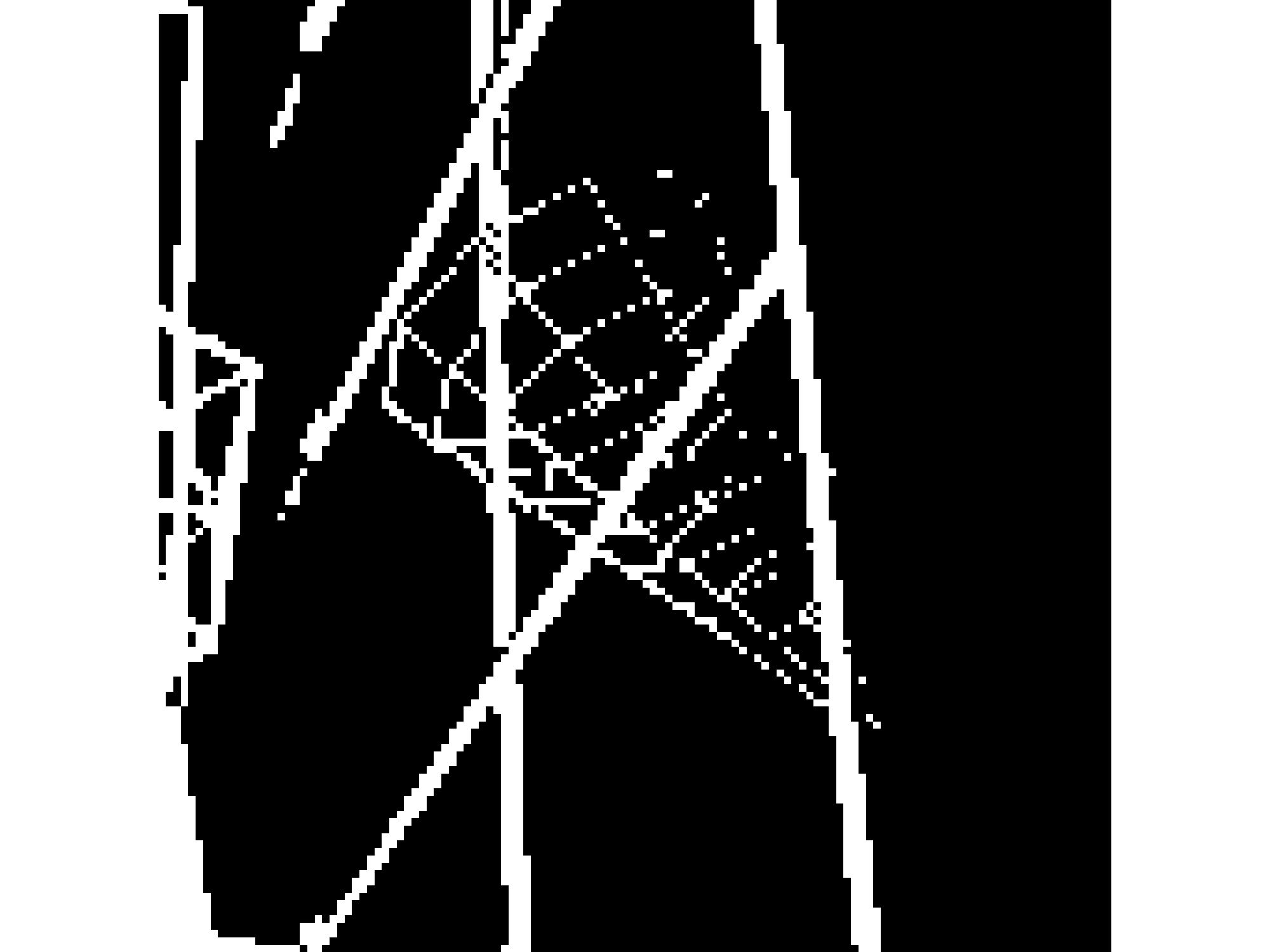}
                                \vspace{-0.35cm}
          \hspace{-2.5cm}    
        \end{subfigure}%
        ~ 
        \begin{subfigure}[b]{0.18\textwidth}
               \hspace{-0.5cm}
       \vspace{0.17cm}
                \includegraphics[width=\textwidth]{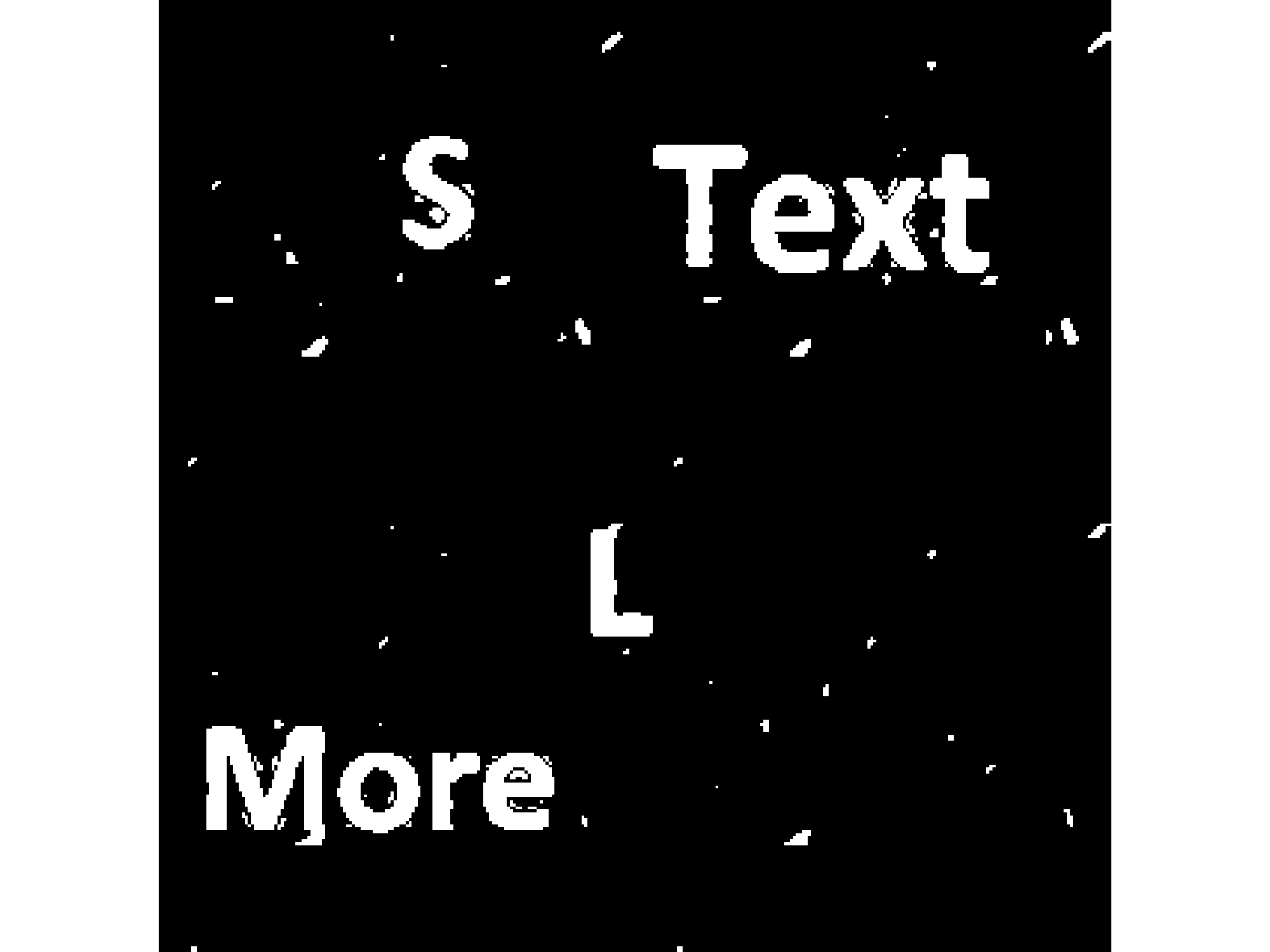}
                \vspace{-0.5cm}
            \hspace{-3cm} 
        \end{subfigure}%
        ~ 
        \begin{subfigure}[b]{0.18\textwidth}
               \hspace{-0.995cm}
       \vspace{0.13cm}
                \includegraphics[width=\textwidth]{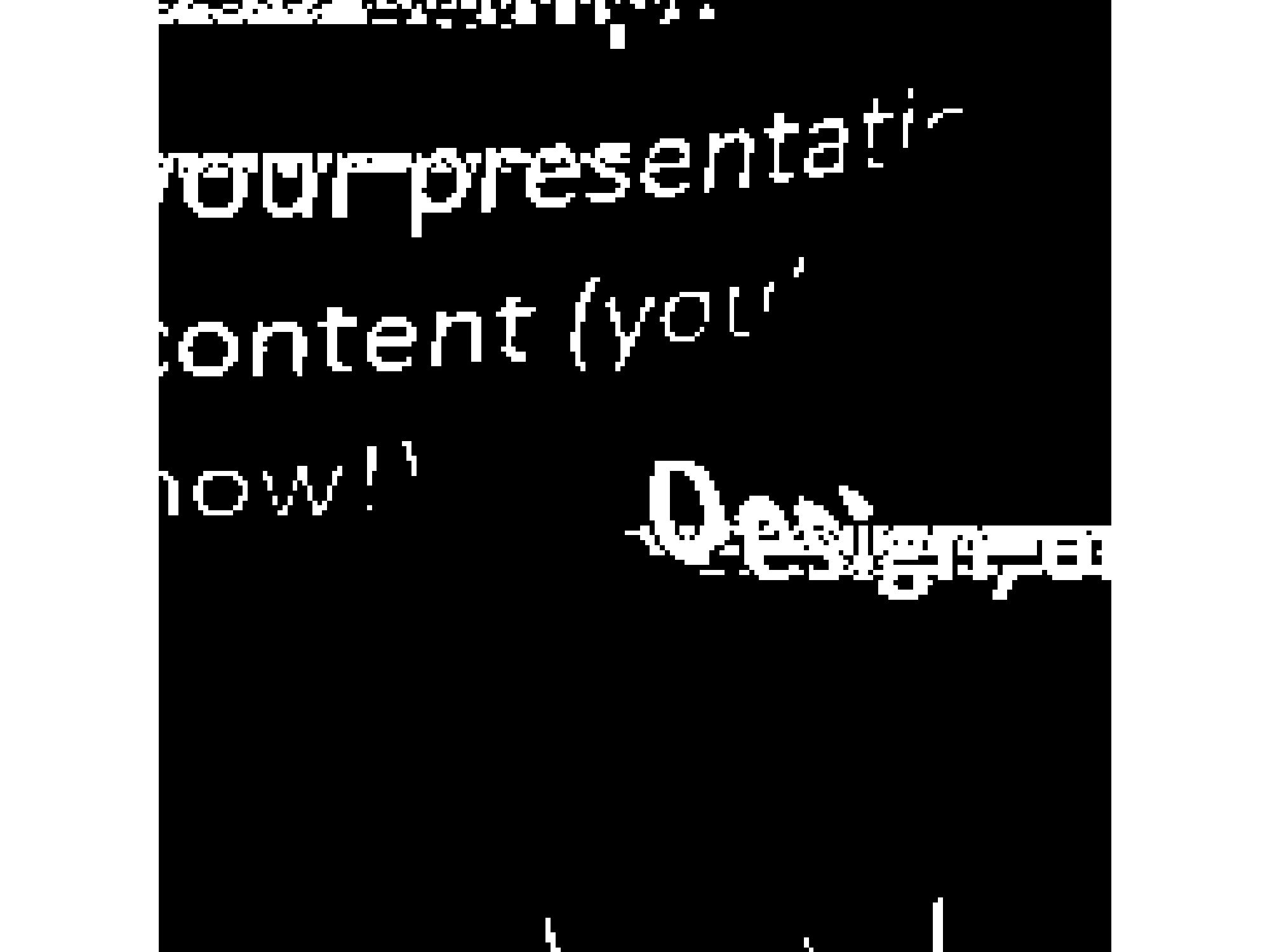}
                 \vspace{-0.45cm}
              \hspace{-4.8cm}
        \end{subfigure}
         \\[1ex]         
                 \centering
        \vspace{-0.05cm}
        \hspace{-1.2cm}
        \begin{subfigure}[b]{0.18\textwidth}
                \includegraphics[width=\textwidth]{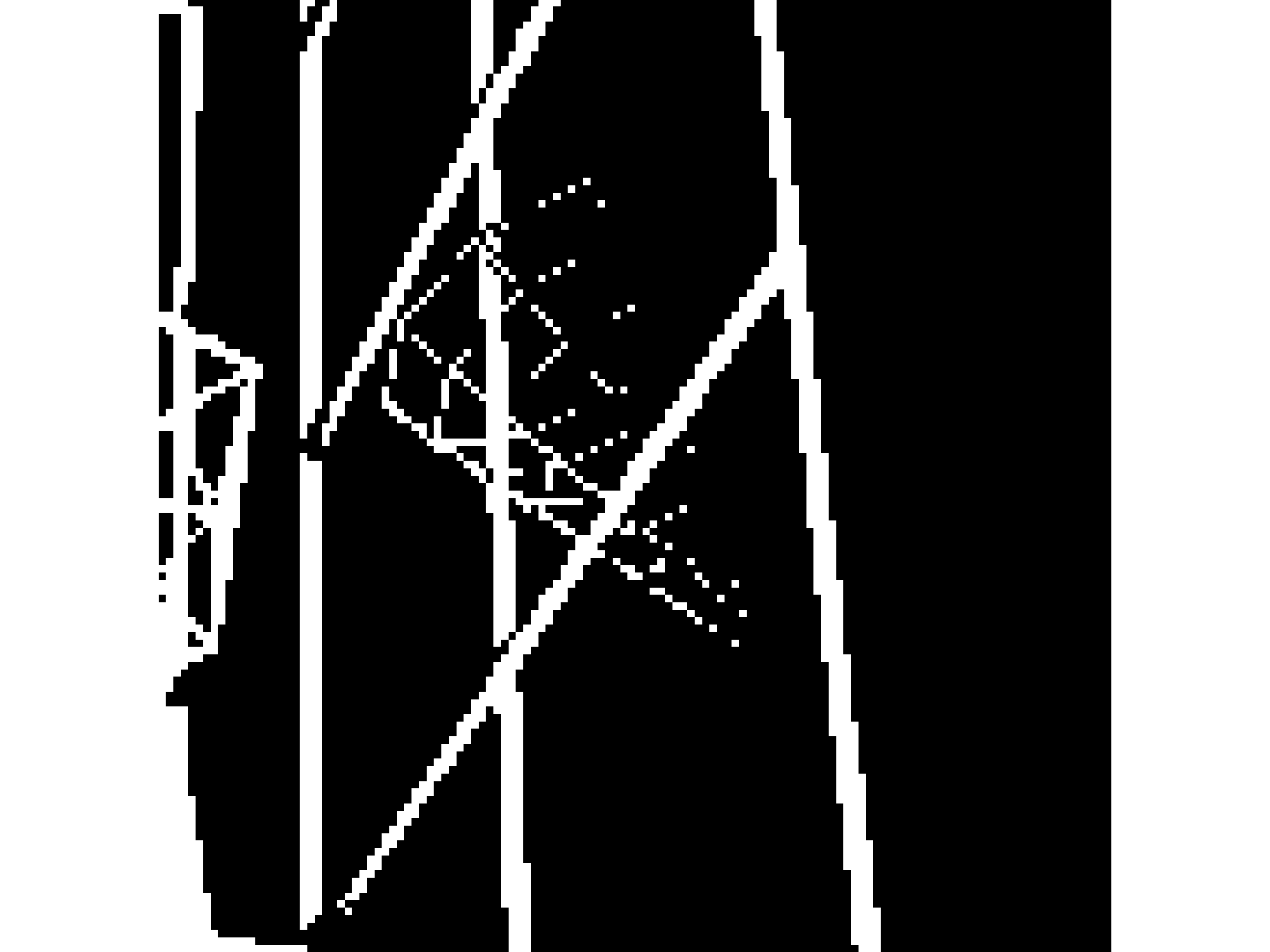}
                                \vspace{-0.35cm}
          \hspace{-2.5cm}    
        \end{subfigure}%
        ~ 
        \begin{subfigure}[b]{0.18\textwidth}
                \hspace{-0.5cm}
       \vspace{0.17cm}
                \includegraphics[width=\textwidth]{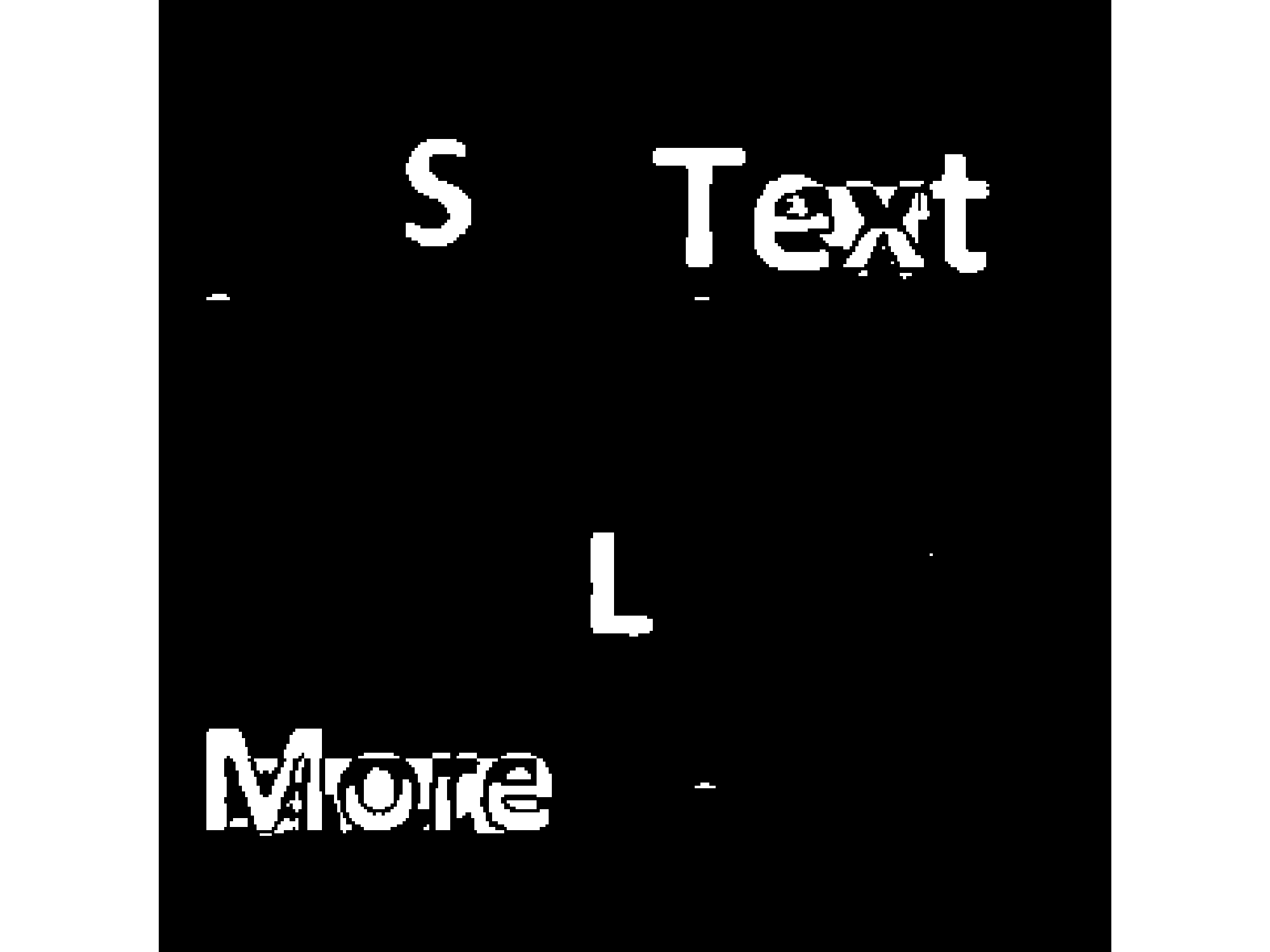}
                \vspace{-0.5cm}
            \hspace{-3cm} 
        \end{subfigure}%
        ~ 
        \begin{subfigure}[b]{0.18\textwidth}
                \hspace{-0.995cm}
       \vspace{0.13cm}
                \includegraphics[width=\textwidth]{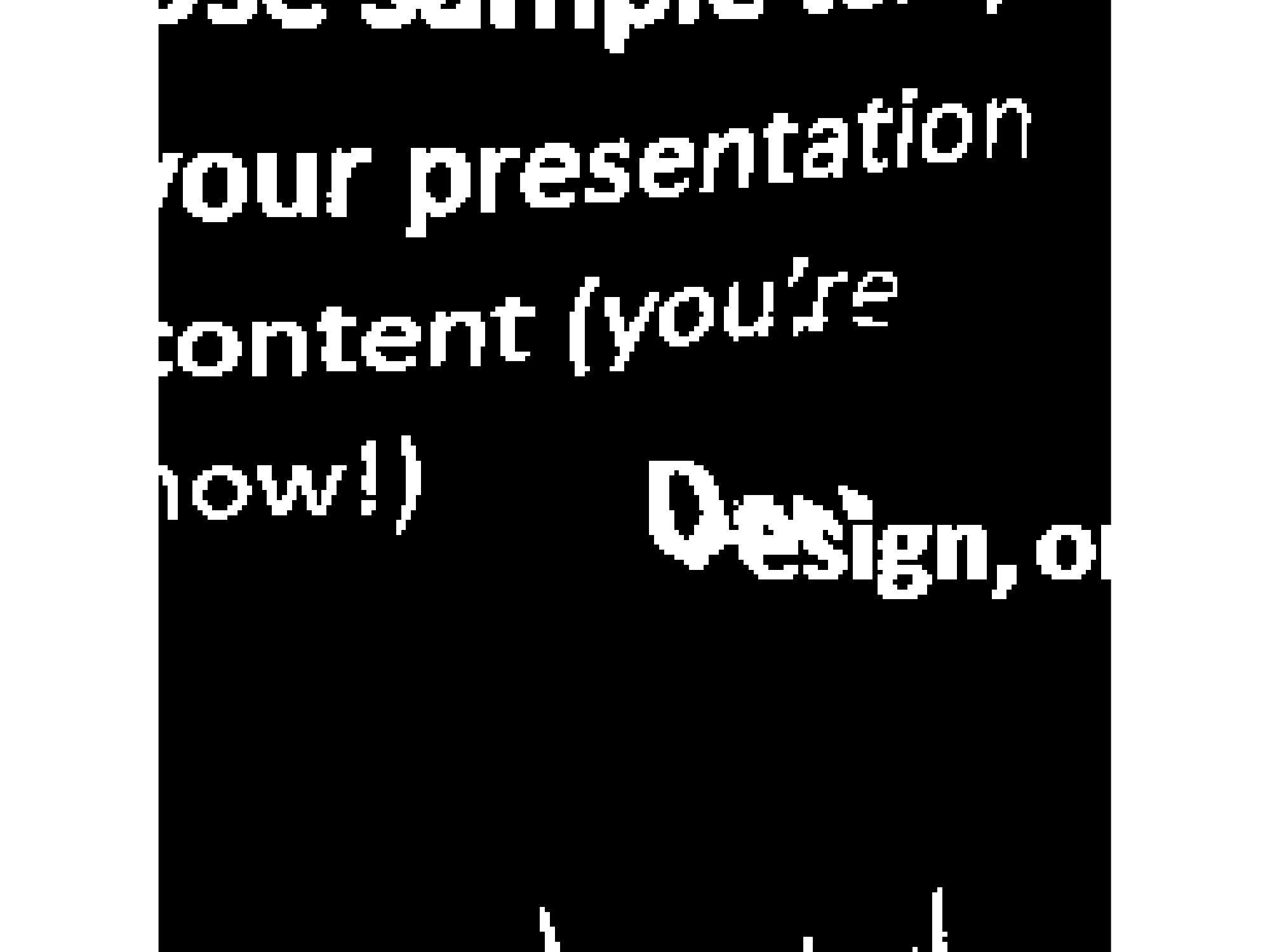}
                 \vspace{-0.45cm}
              \hspace{-4.8cm}
        \end{subfigure}
         \\[1ex]     
                           \centering
        \vspace{-0.05cm}
        \hspace{-1.2cm}
        \begin{subfigure}[b]{0.18\textwidth}
                \includegraphics[width=\textwidth]{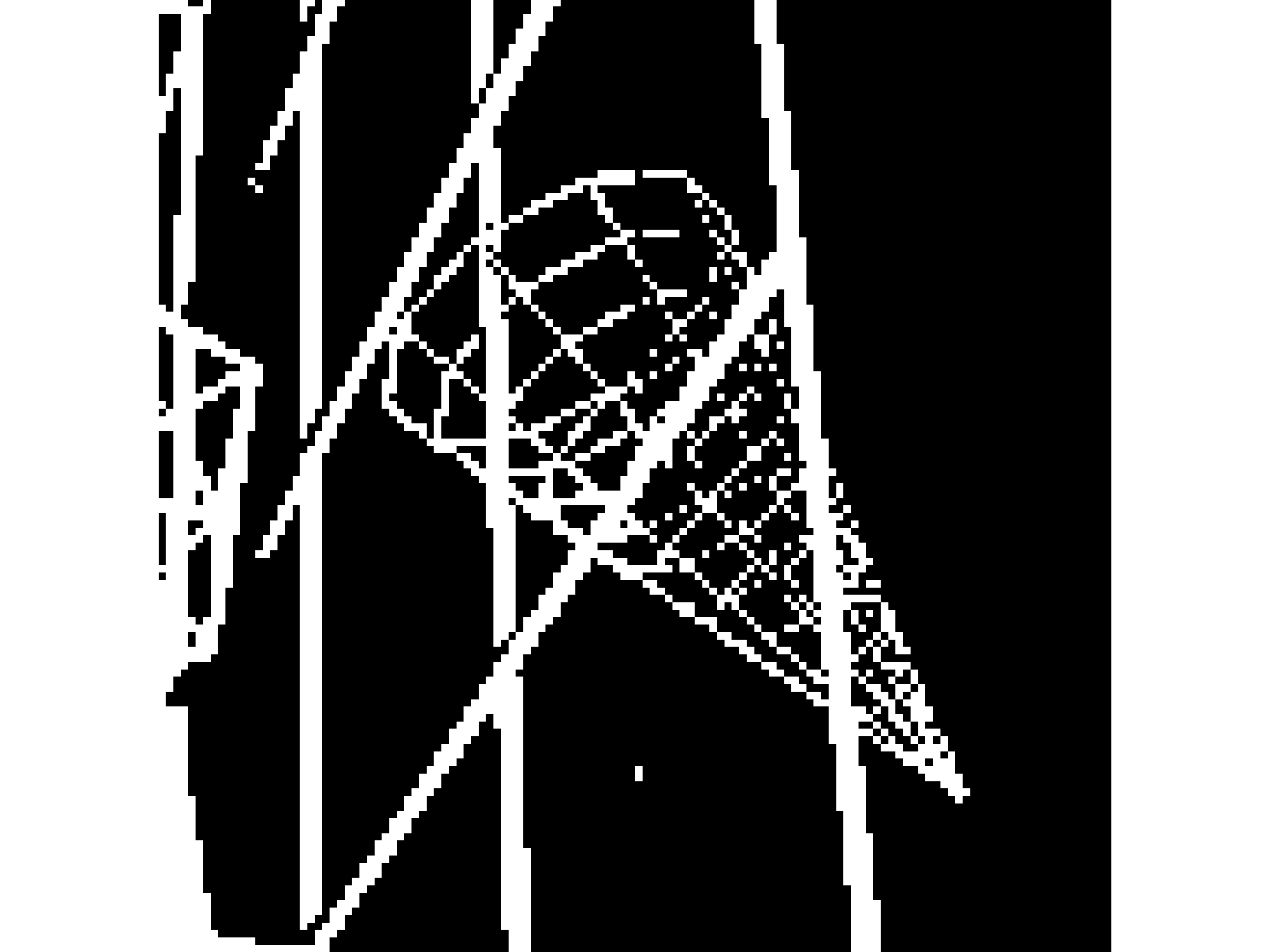}
                                \vspace{-0.1cm}
          \hspace{-2.5cm}    
        \end{subfigure}%
        ~ 
        \begin{subfigure}[b]{0.18\textwidth}
               \hspace{-0.5cm}
       \vspace{0.42cm}
                \includegraphics[width=\textwidth]{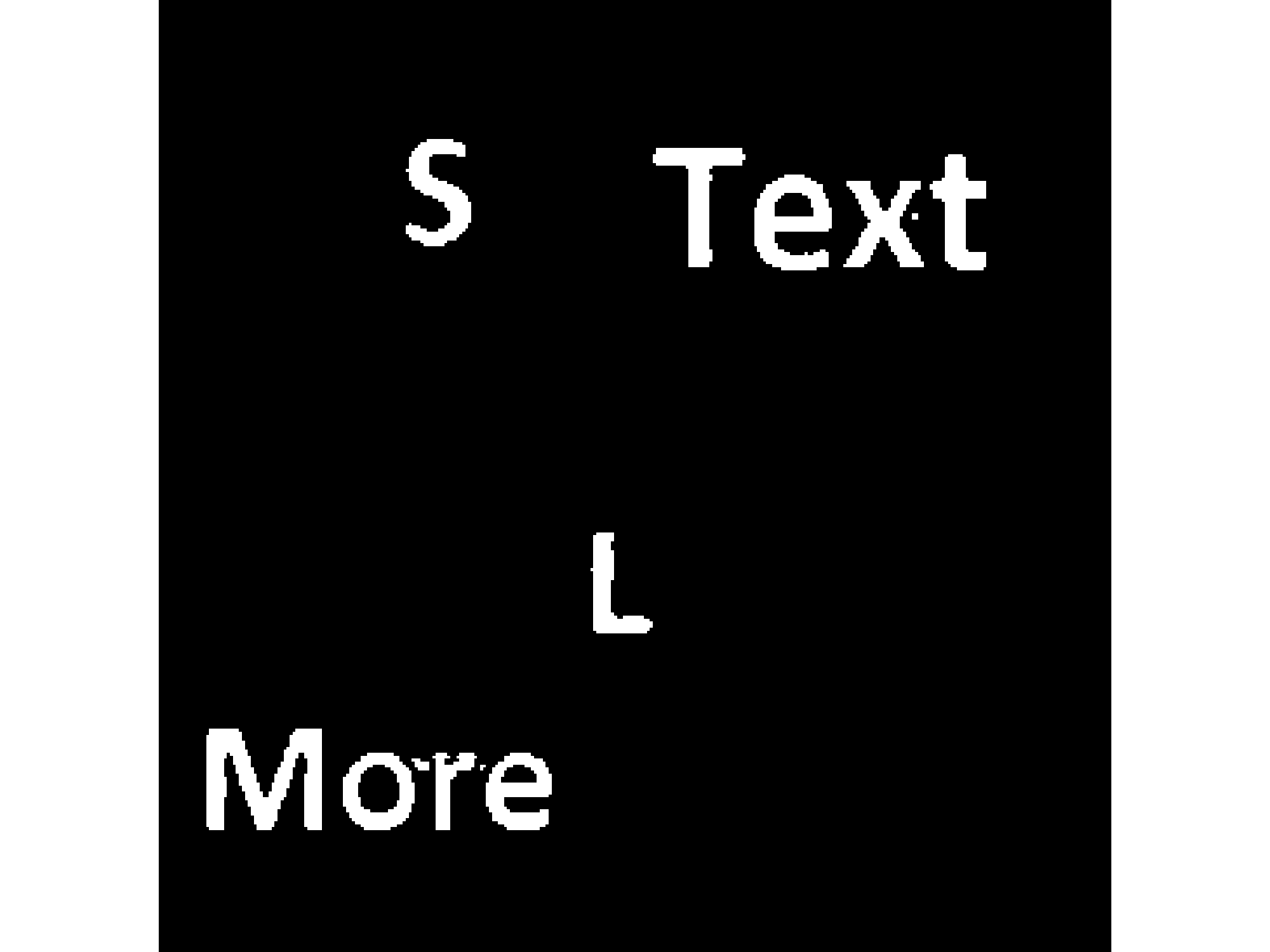}
                \vspace{-0.5cm}
            \hspace{-3cm} 
        \end{subfigure}%
        ~ 
        \begin{subfigure}[b]{0.18\textwidth}
               \hspace{-0.995cm}
       \vspace{0.38cm}
                \includegraphics[width=\textwidth]{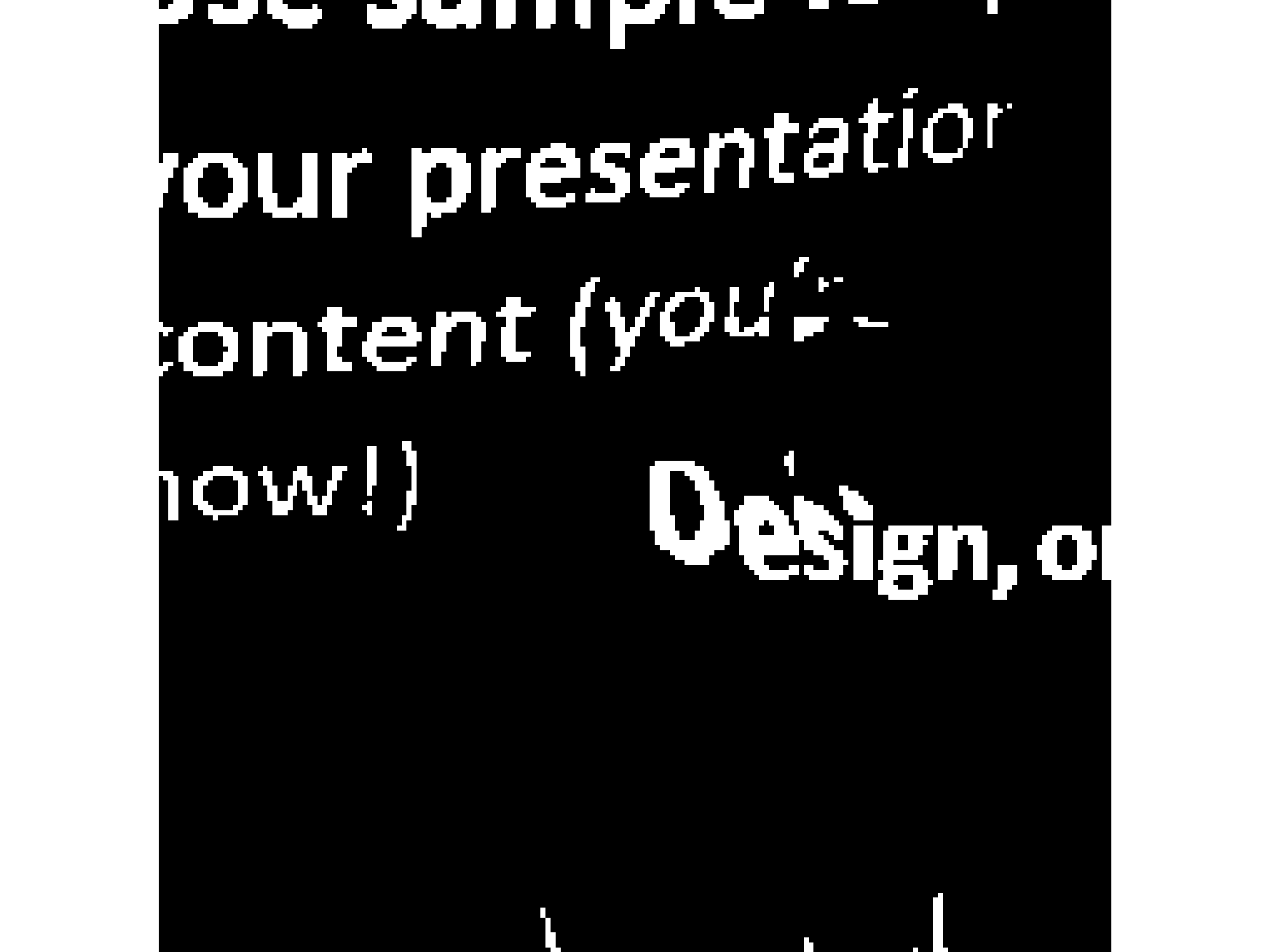}
                 \vspace{-0.45cm}
              \hspace{-4.8cm}
        \end{subfigure}    
        \caption{ Segmentation result for the selected test images. The images in the first to sixth rows denote the original image, and the foreground map by shape primitive extraction and coding, hierarchical k-means clustering, sparse and low-rank decomposition, least absolute deviation fitting and the proposed algorithm respectively.}
\end{figure}

To see the visual quality of the segmentation, the results for 3 test images (each consisting of multiple 64$\times$64 blocks) are shown in Fig. 3. 
It can be seen that the proposed algorithm gives superior performance over DjVu and SPEC in all cases.
There are also noticeable improvement over least absolute deviation (LAD) fitting and low-rank decomposition in one of the images.
We would like to note that, this dataset mainly consists of challenging images where the background and foreground have overlapping color ranges. For simpler cases where the background has a narrow color range that is quite different from the foreground, DjVu, LAD and low-rank decomposition  will work well.

\section{Conclusion}
This paper proposed a subspace learning algorithm for a set of smooth signals in the presence of structured outliers and noise.
The outliers are assumed to be sparse and connected, and suitable regularization terms are added to the optimization framework to promote this properties.
We then solve the optimization problem by alternatively updating the model parameters, and the subspace.
We also show the application of this framework for background-foreground segmentation in still images, where the foreground can be thought as the outliers in our model, and achieve better results than the previous algorithms for background/foreground separation.

\section*{Acknowledgment}
The authors would like to thank Ivan Selesnick, Arian Maleki, and Carlos Fernandez-Granda for their valuable comments and feedback.
We would also like to thanks Stephen Becker for providing the MATLAB implementation of fastRPCA algorithm.


\begin{thebibliography}{1}
\begin{small}

\bibitem{pca}
H Abdi, LJ Williams, ``Principal component analysis'', Wiley Interdisciplinary Reviews: Computational Statistics, 2010.
\bibitem{lda}
AJ Izenman, ``Linear discriminant analysis'', Modern Multivariate Statistical Techniques, Springer, pp. 237-280, 2013.
\bibitem{lpp}
X Niyogi, ``Locality preserving projections'', In Neural information processing systems, vol. 16, 2004.
\bibitem{lsf}
PJ Huber, ''Robust statistics'', Springer Berlin Heidelberg, 2011.
\bibitem{rsl}
F De La Torre, MJ Black, ``A framework for robust subspace learning'', International Journal of Computer Vision,  pp.117-142, 2003.
\bibitem{rpca}
J Wright, A Ganesh, S Rao, Y Peng, Y Ma, ``Robust principal component analysis: Exact recovery of corrupted low-rank matrices via convex optimization'', In Advances in neural information processing systems, 2009.
\bibitem{lerman}
G Lerman, MB McCoy, JA Tropp, T Zhang, ``Robust computation of linear models by convex relaxation'', Foundations of Computational Mathematics 15, no. 2: 363-410, 2015.
\bibitem{grasta}
J He, L Balzano, A Szlam, ``Incremental gradient on the grassmannian for online foreground and background separation in subsampled video'', In Computer Vision and Pattern Recognition, pp. 1568-1575, IEEE, 2012.
\bibitem{tgrasta}
J He, D Zhang, L Balzano, T Tao, ``Iterative online subspace learning for robust image alignment'',  International Conference and Workshops on Automatic Face and Gesture Recognition, IEEE, 2013.
\bibitem{group}
R Chartrand, B Wohlberg, ``A nonconvex ADMM algorithm for group sparsity with sparse groups'', International Conference on Acoustics, Speech and Signal Processing. IEEE, 2013.
\bibitem{tv}
JF Cai, B Dong, S Osher, Z Shen, ``Image restoration: Total variation, wavelet frames, and beyond'', Journal of the American Mathematical Society 25.4: 1033-1089, 2012.
\bibitem{djvu}
P. Haffner, P.G. Howard, P. Simard, Y. Bengio and Y. Lecun, ``High quality document image compression with DjVu'', Journal of Electronic Imaging, 7(3), 410-425, 1998.
\bibitem{spec}
T. Lin and P. Hao, ``Compound image compression for real-time computer screen image transmission'', IEEE Transactions on Image Processing, 14(8), 993-1005, 2005.
\bibitem{LAD}
S. Minaee and Y. Wang, ``Screen content image segmentation using least absolute deviation fitting'', IEEE International Conference on Image Processing, pp.3295-3299, Sept. 2015.
\bibitem{mine1}
J. Zhang and R. Kasturi, ``Extraction of Text Objects in Video Documents:  Recent Progress'', Document Analysis Systems, 2008.
\bibitem{mine2}
S Minaee and Y Wang, ``Screen Content Image Segmentation Using Sparse Decomposition and Total Variation Minimization'', International Conference on Image Processing, IEEE, 2016.
\bibitem{mine4}
S Zhang, Y Zhan, M Dewan, J Huang, DN Metaxas, S Zhou, ``Towards robust and effective shape modeling: Sparse shape composition'', Medical image analysis, 2012.
\bibitem{mine5}
A Taalimi, H Qi, R Khorsandi, ``Online multi-modal task-driven dictionary learning and robust joint sparse representation for visual tracking'', Advanced Video and Signal Based Surveillance (AVSS), IEEE, 2015.
\bibitem{mine6}
S Minaee and Y Wang, ``Screen Content Image Segmentation Using Robust Regression and Sparse Decomposition'', IEEE Journal on Emerging and Selected Topics in Circuits and Systems, 2016.
\bibitem{mine7}
S Minaee, A Abdolrashidi and Y Wang, ``Screen Content Image Segmentation Using Sparse-Smooth Decomposition'', Asilomar Conference on Signals,Systems, and Computers, IEEE, 2015.
\bibitem{lowrank}
A Aravkin, S Becker, V Cevher, P Olsen, ``A variational approach to stable principal component pursuit'', Conference on Uncertainty in Artificial Intelligence, 2014.



\bibitem{mairal}
F Bach, R Jenatton, J Mairal, G Obozinski, ``Optimization with sparsity-inducing penalties''," Foundations and Trends in Machine Learning 4.1: 1-106, 2012.
\bibitem{admm}
S. Boyd, N. Parikh, E. Chu, B. Peleato and J. Eckstein, ``Distributed optimization and statistical learning via the alternating direction method of multipliers'', Foundations and Trends in Machine Learning, 3(1), 1-122, 2011.
\bibitem{prox}
PL Combettes, JC Pesquet, ``Proximal splitting methods in signal processing'', Fixed-point algorithms for inverse problems in science and engineering. Springer, 185-212, 2011.
\bibitem{gram}
SJ Leon, A Bjorck, W Gander,  ``Gram‐Schmidt orthogonalization: 100 years and more'', Numerical Linear Algebra with Applications 20.3: 492-532, 2013.


\bibitem{our_dataset}
https://sites.google.com/site/shervinminaee/research/image-segmentation
\bibitem{becker}
https://github.com/stephenbeckr/fastRPCA
\bibitem{metrics}
DM Powers, ``Evaluation: from precision, recall and F-measure to ROC, informedness, markedness and correlation'', 2011.


\end{small}
\end{thebibliography}
\end{document}